\documentclass[11pt]{article}

\usepackage[final]{acl}

\usepackage{times}
\usepackage{latexsym}

\usepackage[T1]{fontenc}

\usepackage[utf8]{inputenc}

\usepackage{microtype}

\usepackage{inconsolata}

\usepackage{graphicx}

\usepackage{amsmath,amssymb,hyperref}
\usepackage{multirow}
\usepackage{multicol}
\usepackage[export]{adjustbox}
\usepackage{enumitem}
\usepackage{scrextend}

\title{Reusing Latent Speech Representations for \\Query-Conditioned Topic Localization in Transcripts}

\author{
 \textbf{Steffen Freisinger},
 \textbf{Philipp Seeberger},
 \textbf{Thomas Ranzenberger},
 \textbf{Tobias Bocklet},
\\
 \textbf{Korbinian Riedhammer},
\vspace{0.5em}
\\
Technische Hochschule Nürnberg Georg Simon Ohm
\\
\small{
  \textbf{Correspondence:} \href{mailto:steffen.freisinger@th-nuernberg.de}{steffen.freisinger@th-nuernberg.de}
}
}

\begin{document}
\maketitle
\begin{abstract}
Long transcripts are costly inputs for downstream NLP systems and often contain irrelevant context. 
We study query-conditioned topic localization: predicting the sentence span in a transcript that best addresses a topic-title query. 
To improve span localization, we reuse ASR encoder states as sentence-level representations and fuse them with textual embeddings. 
This lets lightweight span locators exploit speech information without running a separate audio encoder. 
Experiments on two public datasets show consistent gains over text-only baselines, especially under strict boundary-matching criteria. 
Cross-dataset experiments further indicate that the benefits are strongest for structured or semi-structured speech, while gains on spontaneous speech are limited and mixed.
\end{abstract}

\section{Introduction}
Long speech transcripts are increasingly used in NLP pipelines for retrieval, question answering, and summarization.
However, processing full transcripts is often inefficient and unreliable.
Standard attention scales quadratically with input length, and long-context models can miss information buried in the middle of a document~\cite{Liu-2024,Levy-2024,Modaressi-2025,Li-2025}.
A common solution is to retrieve shorter passages before downstream processing.
Yet, fixed-size windows ignore topical structure, while topic segmentation produces query-independent segments that may be misaligned with a user's information need.

We study a more targeted setting: \emph{query-conditioned topic localization}.
Given a textual query and a transcript, the goal is to predict the contiguous sentence span that best addresses the query.
This task is closely related to span localization in videos and meetings, where models such as VSLNet and QMSum-style locators predict start and end positions conditioned on a query~\cite{Zhang-2020,Zhong-2021,Yang-2023}.
Most approaches, however, rely mainly on text and vision input.
This discards acoustic cues such as pauses, prosody, speaker changes, background music, and scene transitions, which can mark topic boundaries and have been shown to help spoken topic segmentation~\cite{Ghinassi-2023,Freisinger-2026,Retkowski-2026}.
At the same time, spoken documents are usually processed by automatic speech recognition (ASR).
This raises a simple question: can we reuse the internal representations of the ASR model to improve query-conditioned localization?

We propose an audio-augmented localization framework that reuses ASR encoder states.
For each sentence, we pool the corresponding encoder states into an ASR-derived representation and fuse it with a textual sentence embedding.
The resulting audio-text representation can be used by lightweight span locators without running a separate audio encoder.
We apply this idea to two model families: (i) QA-based architectures (i.e., VSLNet) and (ii) pointer-based models (QMSum)~\cite{Zhang-2020,Zhong-2021}.
In both cases, the localization model itself remains unchanged; only the input representation is augmented.
Our results indicate that the benefits are strongest for structured and semi-structured spoken content, while gains on spontaneous meetings are weak or mixed.

Our \textbf{contributions} are:
\begin{itemize}[leftmargin=*]
    \itemsep0em
    \item \textbf{Method}: We propose an audio-text framework for query-conditioned topic localization that reuses ASR encoder states as sentence-level representations.
    \item \textbf{Experiments}: We evaluate the approach with two span locators, two ASR models, and four public datasets across different domains and languages.
    \item \textbf{Analysis}: We analyze encoder layers, pooling and fusion strategies, cross-dataset transfer, and qualitatively inspect acoustic boundary cues.
\end{itemize}

\begin{figure*}[t]
  \centering
  \includegraphics[width=0.83\linewidth]{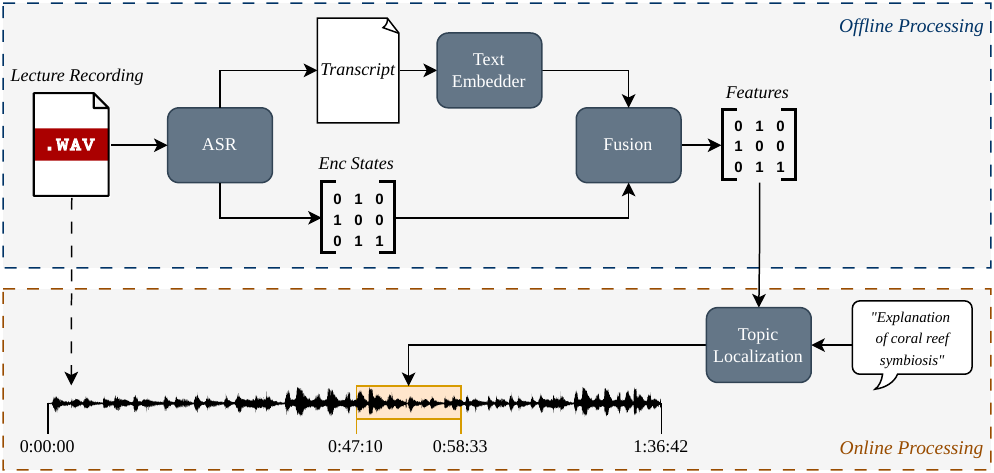}
  \caption{Visualization of the topic localization task and our audio-augmented framework.}
  \label{fig:overview}
\end{figure*}

\section{Related Work}
\label{sec:related}

\subsection{Query-Conditioned Span Localization}
Query-conditioned span localization aims to predict a start and end position for a given query.
In natural language video localization, models ground a text query in a video moment.
\citet{Zhang-2020} propose VSLNet, which treats this task as span-based question answering with context-query attention and query-guided highlighting.
Later work extends this idea to long videos and chapter grounding, where a chapter title or text query is localized in video content~\citep{Zhang-2021,Yang-2023}.
For meetings, \citet{Zhong-2021} introduce QMSum, a locate-then-summarize pipeline in which a pointer-style locator first selects a query-relevant transcript span.
Our pointer model follows this answer-pointer line of work, where start and end positions are predicted with query-conditioned attention~\citep{Wang-2017,Vinyals-2015}.
Unlike passage retrieval, these methods do not only rank candidate segments, but predict fine-grained span boundaries.
Structurally, our task follows the same query-conditioned start/end prediction formulation.
We use the term \emph{topic localization} because the target spans correspond to topics and topic titles are used as queries.
Our contribution lies in augmenting this established localization formulation with representations derived from ASR encoder states.
However, prior localization work is mainly text- or vision-based and does not study acoustic cues for localizing spans in speech transcripts.

\subsection{ASR Representations for Downstream Tasks}
Recent work shows that ASR encoders can be reused for tasks beyond transcription.
\citet{Gong-2023} show that the Whisper encoder can be frozen and combined with a lightweight classifier for audio tagging.
\citet{Wang-2023} transfer Whisper to spoken language understanding without using a separate audio encoder.
\citet{Yang-2021} further show in SUPERB that pretrained speech encoders provide useful representations for many downstream speech tasks.
\citet{Li-2024} exploit intermediate Whisper representations for keyword spotting and contextual biasing.
Layer-wise analyses also suggest that different encoder depths capture different acoustic and linguistic information~\citep{Pasad-2021,Baumann-2023}.
We build on this reuse paradigm, but apply ASR encoder states to sentence-level, query-conditioned span localization.

\subsection{Audio Cues for Spoken Topic Segmentation}
Topic segmentation divides a document into query-independent segments based on thematic shifts.
It has been studied with unsupervised and supervised text-based methods~\citep{Hearst-1997,Koshorek-2018,Lukasik-2020}.
Because automatic transcripts often lack structure, topic segmentation is also widely used for spoken content~\citep{Soares-2018,Retkowski-2024,Freisinger-2025}.
Several methods incorporate audio cues.
\citet{Zhang-2019} show that speaker information can improve dialogue segmentation.
More recent approaches use pretrained audio encoders to derive acoustic features for spoken topic segmentation~\citep{berlage-2020,Ghinassi-2023,Freisinger-2026,Retkowski-2026}.
These works show that audio can help detect topical boundaries.
Our work differs in two ways: we study query-conditioned localization rather than query-independent segmentation, and we derive auxiliary representations from precomputed ASR encoder states instead of using a separate audio model.

\section{Methods}
\label{sec:methods}
In this section, we formally define the task of topic localization in spoken content and then introduce our proposed audio-augmented framework, separated into an \emph{offline processing} stage (indexing of recordings) and an \emph{online processing} stage (prediction of the query-relevant span).

\subsection{Task Definition}
\label{ssec:definition}
We define topic localization as query-conditioned span prediction over a speech transcript.
Let a transcript be split into $L$ sentences,
$\mathbf{d}=(s_1,\ldots,s_L)$, and let $q$ be a textual query.
The goal is to predict one contiguous sentence span
\[
(i^\star,j^\star),\qquad 1\le i^\star \le j^\star \le L,
\]
such that $(s_{i^\star},\ldots,s_{j^\star})$ best addresses $q$.
We treat the transcript sentences as the context and the topic title as the query.
This provides a controlled, annotation-aligned query setting rather than unrestricted natural user queries.
We examine sensitivity to alternative query formulations in Appendix~\ref{app:query-perturbation}.
The model always outputs a single span.
\autoref{fig:overview} gives an overview of the task and framework.

\subsection{Offline: Sentence-Level Context Encoding}
\label{ssec:offlineprocessing}

We model a pipeline in which spoken documents are transcribed once and prepared for later query-time localization.
Audio is segmented with voice activity detection and chunked to $\leq$30\,s for ASR.
The ASR model produces the transcript and encoder states.
The transcript is split into sentences using Punkt~\cite{Kiss-2006}, and the encoder states are kept as a byproduct for acoustic sentence encoding.

\paragraph{Text sentence embeddings.}
Given a transcript with $L$ sentences, we encode each sentence $s_\ell$ with a pretrained text embedding model.
This yields a textual embedding matrix $\mathbf{T}\in\mathbb{R}^{L\times d_{\text{text}}}$.

\paragraph{ASR-derived audio sentence embeddings.}
For each sentence $s_\ell$, we use its aligned time interval to select the corresponding ASR encoder frames from layer $k$:
\[
\mathbf{E}^{(k)}_\ell =
[\mathbf{e}^{(k)}_{\ell,1},\dots,\mathbf{e}^{(k)}_{\ell,N_\ell}]
\in \mathbb{R}^{N_\ell \times d_{\text{enc}}},
\]
where $N_\ell$ is the number of aligned frames.
A temporal pooling function $g(\cdot)$ maps this frame sequence to one ASR-derived audio sentence vector,
\[
\mathbf{a}^{(k)}_\ell = g(\mathbf{E}^{(k)}_\ell)
\in\mathbb{R}^{d_{\text{aud}}}.
\]
We compare four pooling variants:

\begin{itemize}[leftmargin=*]
    \item \textbf{Mean}: average all frames within the sentence interval,
    $\mathbf{a}^{(k)}_\ell =
    \frac{1}{N_\ell}\sum_{t=1}^{N_\ell}\mathbf{e}^{(k)}_{\ell,t}$.

    \item \textbf{Mean$^{\text{ext}}$}: extend the sentence interval on both sides and average the enlarged frame sequence.
    The extension covers $1.28$s for all ASR models.
    This includes acoustic context around sentence boundaries, where topic-shift cues such as pauses, speaker changes, or sound events may occur~\cite{Freisinger-2026}.

    \item \textbf{Mean+Std}: concatenate the temporal mean with the per-dimension standard deviation, following prior work on audio-based topic segmentation~\cite{Ghinassi-2023}.

    \item \textbf{Boundary}: average two fixed-size windows around the sentence start and end.
    Each window spans $2.56$s.
    The two window means $\mathbf{b}^{(k)}_{\ell,\text{start}}$ and $\mathbf{b}^{(k)}_{\ell,\text{end}}$ are fused as
    $\mathbf{a}^{(k)}_\ell =
    \tfrac{1}{2}(\mathbf{b}^{(k)}_{\ell,\text{start}}+
    \mathbf{b}^{(k)}_{\ell,\text{end}})$.
\end{itemize}

Applying the pooling to all sentences yields an audio embedding matrix
$\mathbf{A}\in\mathbb{R}^{L\times d_{\text{aud}}}$.

\begin{figure}[t]
  \centering 
  \includegraphics[width=\columnwidth]{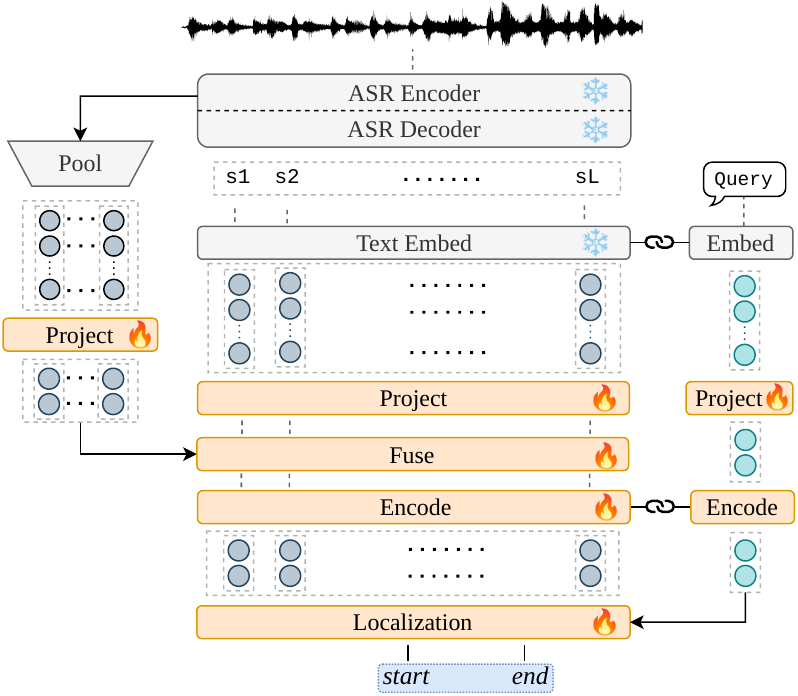} 
  \caption{Architecture overview of our audio-augmented topic localization setup. 
    \includegraphics[height=1em, valign=c]{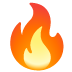}~means that model weights are updated during training, 
    \includegraphics[height=1em, valign=c]{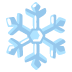}~indicates frozen weights, 
    \includegraphics[height=1em, valign=c]{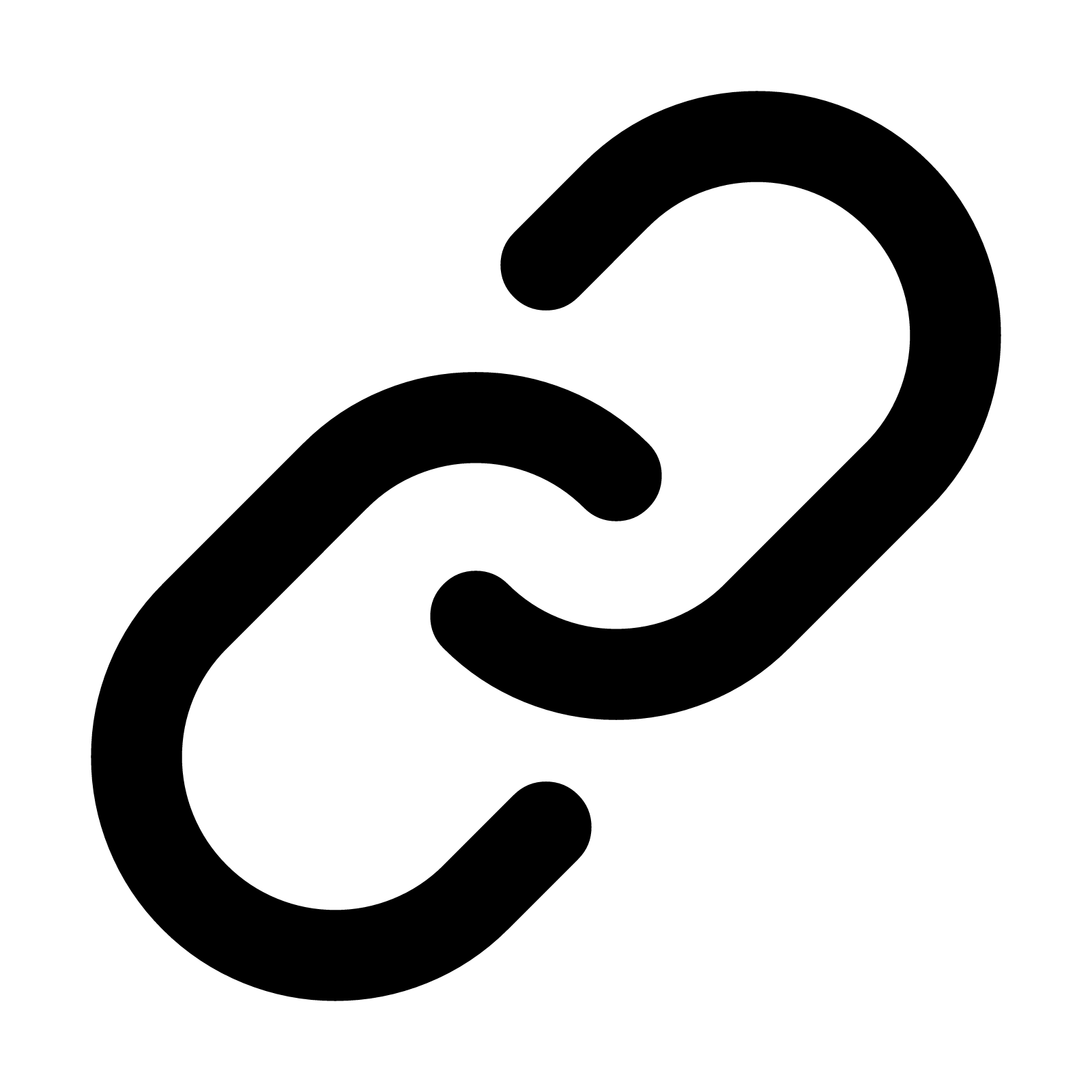}~indicates shared model weights.} 
  \label{fig:architecture}
\end{figure}

\paragraph{Projection and fusion.}
We project text and audio embeddings to a shared hidden size $d$:
\[
\tilde{\mathbf{T}}=\mathbf{T}\mathbf{W}_{\text{text}},
\qquad
\tilde{\mathbf{A}}=\mathbf{A}\mathbf{W}_{\text{aud}},
\]
with $\tilde{\mathbf{T}},\tilde{\mathbf{A}}\in\mathbb{R}^{L\times d}$.
For multimodal input, we compare three sentence-level fusion operators:
\begin{itemize}[leftmargin=*]
    \item \textbf{Sum}: $\mathbf{V}=\tilde{\mathbf{T}}+\tilde{\mathbf{A}}$.
    \item \textbf{Concat}: $\mathbf{V}=\phi([\tilde{\mathbf{T}};\tilde{\mathbf{A}}])$, where $\phi:\mathbb{R}^{2d}\to\mathbb{R}^{d}$ is a linear layer with ReLU and dropout.
    \item \textbf{Gated}: $\mathbf{V}=\mathbf{g}\odot\tilde{\mathbf{T}}+(1-\mathbf{g})\odot\tilde{\mathbf{A}}$, where
    $\mathbf{g}=\sigma(G([\tilde{\mathbf{T}};\tilde{\mathbf{A}}]))$ and $G$ is a two-layer MLP.
    This follows the idea of gated multimodal fusion~\cite{Arevalo-2017}.
\end{itemize}
Depending on the input setting, the context representation is either $\tilde{\mathbf{T}}$ (text-only), $\tilde{\mathbf{A}}$ (audio-only), or $\mathbf{V}$ (text+audio).

\paragraph{Context encoding.}
The selected context representation is passed through a QANet-style encoder with convolutions and multi-head self-attention~\cite{Yu-2018}.
This yields the final context matrix $\mathbf{C}\in\mathbb{R}^{L\times d}$.
Since $\mathbf{C}$ is independent of the query, it is computed once per recording and reused during online localization.
Implementation details are given in Section~\ref{ssec:expsetup}.

\subsection{Online: Query-Conditioned Localization}
At query time, the precomputed context matrix $\mathbf{C}$ is combined with an encoded query to predict the query-relevant sentence span.

\paragraph{Query encoding.}
We encode the textual query with the same pretrained text embedding model used for the transcript.
Unlike the context sentences, the query is represented at token level:
$\mathbf{Q}\in\mathbb{R}^{J\times d_{\text{text}}}$, where $J$ is the number of query tokens.
We project $\mathbf{Q}$ to the hidden size $d$ and pass it through the same QANet-style encoder used for the context to obtain
$\tilde{\mathbf{Q}}\in\mathbb{R}^{J\times d}$.
The precomputed context matrix $\mathbf{C}$ and encoded query $\tilde{\mathbf{Q}}$ are then passed to one of two span locators.
Both locators retain their established query-conditioned start/end prediction formulation.
Our framework changes the input representation used by these models.

\paragraph{VSLNet.}
We adapt VSLNet to sentence-level topic localization~\cite{Zhang-2020,Zhang-2021}.
The context matrix $\mathbf{C}$ is treated as the passage and the encoded query $\tilde{\mathbf{Q}}$ as the question.
A context-query attention module aligns both inputs, and a span head predicts start and end distributions over sentence positions.
We keep the original architecture and losses, including query-guided highlighting; only the input representation is changed.

\paragraph{QMSum-Pointer.}
We also adapt the locator from QMSum~\cite{Zhong-2021}.
A start pointer scores all sentence positions conditioned on the query.
An end pointer then predicts the end position conditioned on the query and the selected start representation.
At inference, we choose the highest-scoring valid span $(i,j)$ with $i\le j$.


\section{Experiments}
\label{sec:experiments}
We evaluate whether ASR encoder states provide useful auxiliary information for query-conditioned topic localization.
We first test our models on two main datasets, then analyze design choices and cross-dataset transfer.
We release the evaluation code and checkpoints for our localization models\footnote{\url{https://github.com/steffrs/speech-topic-localization}\label{fnrepo}}.

\begin{table*}[t]
  \centering
  \renewcommand{\arraystretch}{1.1}
  \small
  \begin{tabular}{ll|rr|cccc|cccc}
    \hline
    \multirow{3}{*}{\textbf{Method}} & \multirow{3}{*}{\textbf{Input}} &  &  & \multicolumn{4}{c|}{\textsc{Euronews}} & \multicolumn{4}{c}{\textsc{YTSeg}} \\
    \cline{5-12}
     &  &
    \multicolumn{2}{c|}{\textbf{FLOPs}} & 
    \multirow{2}{*}{\textbf{EM}} & \multicolumn{3}{c|}{\textbf{R@1, IoU$\geq$}} & 
    \multirow{2}{*}{\textbf{EM}} & \multicolumn{3}{c}{\textbf{R@1, IoU$\geq$}} \\
     &  & \textit{Offline} & \textit{Online} &  & \textit{0.7} & \textit{0.5} & \textit{0.3} & 
      & \textit{0.7} & \textit{0.5} & \textit{0.3} \\
    \hline
    \multirow{3}{*}{VSLNet} 
            & T   & 126M & \multirow{3}{*}{37.2M} & 45.26 & 62.51 & 72.92 & 81.95 & 13.88 & 32.39 & 44.69 & 56.85 \\  
            & A   & 203M &  & 58.28 & 62.15 & 67.76 & 72.25 & 5.97 & 10.83 & 16.13 & 24.22 \\  
            & T+A & 220M &  & \textbf{69.11} & \textbf{74.09} & \textbf{79.66} & \textbf{84.19} & \textbf{20.56} & \textbf{37.45} & \textbf{48.63} & \textbf{60.50} \\ 
    \hline
    \multirow{3}{*}{\begin{tabular}[c]{@{}c@{}}QMSum-\\Pointer\end{tabular}} 
            & T   & 126M & \multirow{3}{*}{21.2M} & 35.34 & 51.15 & 63.18 & 74.81 & 7.63 & 17.33 & 25.81 & 37.47 \\  
            & A   & 203M &  & 53.93 & 58.42 & 66.05 & 75.53 & 4.43 & 7.95 & 13.39 & 22.65 \\ %
            & T+A & 220M &  & \textbf{61.97} & \textbf{66.91} & \textbf{74.49} & \textbf{81.95} & \textbf{13.10} & \textbf{21.31} & \textbf{28.71} & \textbf{40.31} \\  %
    \hline
  \end{tabular}
  \caption{Main results on \textsc{Euronews} and \textsc{YTSeg}. 
    T=Text embeddings, A=ASR-derived embeddings, T+A=gated fusion. 
    FLOPs are averaged over \textsc{Euronews} test samples and cover localization-stage computation, excluding shared ASR transcription and text-embedding extraction.
    Bold marks the best score for each model and dataset.}
  \label{tab:main-results}
\end{table*}

\subsection{Datasets}
\label{ssec:datasets}
We use two datasets for training and in-domain evaluation: \textsc{Euronews}~\cite{Shukla-2024} and \textsc{YTSeg}~\cite{Retkowski-2024}.
For cross-dataset evaluation, we use two additional datasets without further training: \textsc{AMI}~\cite{AMI-2005} and \textsc{Videoaula}~\cite{Soares-2018}.
All datasets contain speech recordings with topic-segment annotations.
Each segment is defined by a start time, an end time, and a topic title.
We use the title as the query and map the time interval to start and end sentence indices using ASR sentence timestamps.
Thus, each example consists of a transcript, a query, and one target sentence span.
Dataset statistics are shown in \autoref{tab:datastats}.

\paragraph{\textsc{Euronews}}
is based on the \emph{Latest news bulletin} YouTube playlists of \emph{Euronews} and provides chapter timestamps for topical shifts~\cite{Shukla-2024}.
The dataset covers six languages: English, French, German, Italian, Portuguese, and Spanish.
We use the official train/dev/test split.
Since the released annotations do not include topic titles, we extract chapter titles from the timestamped entries in the YouTube video descriptions.
We release the extraction script to make this step reproducible\footref{fnrepo}.

\paragraph{\textsc{YTSeg}}
consists of 19{,}299 English YouTube videos from 393 channels, including formats such as podcasts, lectures, and news. 
The videos cover diverse topics, e.g., science, lifestyle, politics, health, economy, and technology~\cite{Retkowski-2024}. 
We use the validation set for early stopping during training and report results on the test set.

\paragraph{\textsc{AMI}} 
comprises 171 English meeting recordings with 138 sessions annotated for topic segmentation~\cite{AMI-2005}. 
Each recording covers multiple speakers with overlapping and spontaneous speech. 

\paragraph{\textsc{Videoaula}}
contains 34 Portuguese video lectures on computer science~\cite{Soares-2018}. 
The lectures are given by five different speakers.

\subsection{Experimental Setup}
\label{ssec:expsetup}

\paragraph{ASR and text encoders.}
We compare two ASR models for transcription and acoustic sentence encoding.
For Whisper, we use \emph{large-v3}, an encoder-decoder Transformer trained on about 680k hours of multilingual speech data, and obtain word-level timestamps with WhisperX~\citep{Radford-2022,Bain-2023}.
For Canary, we use \emph{canary-1b-v2}, a FastConformer-Transformer ASR model with word-level timestamps and support for 25 European languages~\citep{Sekoyan-2025}.
Their encoder hidden sizes are $d_{\text{enc}}=1280$ and $d_{\text{enc}}=1024$, respectively.
For the main experiments, we use \emph{paraphrase-multilingual-MiniLM-L12-v2}, as it supports multiple languages while remaining compact enough for our lightweight localization setup~\citep{Reimers-2020}.
In an additional text-encoder ablation, we also evaluate \emph{multilingual-e5-large}~\citep{Wang-2024} and \emph{Qwen3-Embedding-4B}~\citep{Zhang-2025}.
All ASR and text embedding models are frozen.

\paragraph{Localization models.}
All trainable localization components use hidden size $d=128$ and dropout $0.1$.
The QANet-style context and query encoders use one encoder block with four convolution layers, kernel size $7$, and $8$ attention heads.
For VSLNet, we use query-guided highlighting with extension ratio $\alpha=0.1$ and optimize start/end cross-entropy plus the highlighting loss.
For QMSum-Pointer, we optimize start/end cross-entropy.
Only the projection, fusion, context/query encoding, and localization layers are trained.
For each input setting, i.e., text-only, audio-only, and text+audio, we train a separate model.

\paragraph{Training.}
We train all models with \texttt{AdamW}, weight decay $0.01$, and an initial learning rate of $3{\times}10^{-4}$.
We use a cosine learning-rate schedule with linear warmup ratio $0.05$ and batch size $32$.
Training runs for at most $20$ epochs with early stopping on the dev set and patience $2$.
All models are implemented in PyTorch and trained on GPUs. 

\paragraph{Evaluation.}
We evaluate single-span topic localization with Exact Match (EM) and R@1 at IoU thresholds $\tau\in\{0.3,0.5,0.7\}$.
EM measures exact boundary accuracy, while R@1 measures whether the predicted span overlaps the gold span under different boundary strictness levels.
We also report average localization-stage FLOPs for the main results, split into offline and online computation.
Offline FLOPs cover representation pooling, projection, fusion, and context encoding; online FLOPs cover query encoding and span localization.
ASR transcription and text-embedding extraction are excluded because they are shared preprocessing steps that can be performed once during offline indexing.

\subsection{Main Results}
\label{ssec:expmain}
\autoref{tab:main-results} reports results on \textsc{Euronews} and \textsc{YTSeg} for the localizers VSLNet and QMSum-Pointer.
For the input representation, we use T for textual sentence embeddings, A for audio sentence embeddings pooled from ASR encoder states, and T+A for their fused representation.

\paragraph{Text+audio improves localization.}
Across both datasets and both localization models, T+A gives the best results.
T+A improves over text-only for both localizers, with larger gains on \textsc{Euronews} than on \textsc{YTSeg}.
The same trend holds for all R@1 thresholds.
This suggests that ASR encoder states provide information that complements textual sentence embeddings.

\paragraph{Gains are strongest under strict boundary criteria.}
The largest relative gains appear for EM and R@1 at IoU$\geq0.7$.
The gains become smaller as the IoU threshold is relaxed.
This pattern suggests that ASR-derived representations are particularly beneficial for precise boundary localization.
They appear to help the model identify the exact start and end positions rather than only the broader query-relevant region.
We analyze this further in \autoref{ssec:embedding-analysis}.

\paragraph{Audio-only performance depends on the dataset.}
On \textsc{YTSeg}, text-only clearly outperforms audio-only.
On \textsc{Euronews}, however, audio-only is competitive and often stronger than text-only.
This may partly reflect the structured production style of news videos, where topic changes often coincide with speaker changes, scene changes, music, or background transitions.
It may also reflect that late ASR encoder layers encode lexical or semantic information useful for localization.

\paragraph{Computation.}
Text+audio increases offline FLOPs because ASR encoder states must be pooled and fused with text representations.
However, for a given locator, T and T+A have the same query-time computation and stored context size, since both are reduced to a precomputed context matrix \(C \in \mathbb{R}^{L \times d}\).
Under our target offline-indexing setup, recordings are processed once and queried many times.
QMSum-Pointer is cheaper online, while VSLNet gives stronger accuracy.

\subsection{Model Ablations}
\label{ssec:ablations}

\begin{table}[t]
  \centering
  \small
  \renewcommand{\arraystretch}{1.1}
  \begin{tabular}{lll|cc} %
    \hline
    \textbf{Encoder} & \textbf{Pooling} & \textbf{Fusion} & \textbf{EM} & \textbf{R@1} \\
    \hline
    \multirow{7}{*}{Whisper} & - & - & 45.26 & 72.92 \\  
      \cline{2-5}
      & \multirow{3}{*}{mean$^{\text{ext}}$} 
                           & gated  & \textbf{69.11} & \textbf{79.66} \\
      &                    & concat & 68.52 & 78.09 \\
      &                    & sum    & 67.89 & 79.07  \\
      \cline{2-5}
      & mean          & gated  & 68.21  &  77.73 \\
      & mean+std   & gated  & 67.80  &  78.40 \\
      & boundary    & gated  & 68.48  &  78.00  \\
    \hline
    \multirow{7}{*}{Canary} & - & - & 46.50 & 73.90 \\  
      \cline{2-5}
      & \multirow{3}{*}{mean$^{\text{ext}}$} 
                           & gated  & 64.73 & 78.39 \\
      &                    & concat & 63.70 & 76.42 \\
      &                    & sum    & 63.25 & 77.18 \\
      \cline{2-5}
      & mean       & gated  & 63.88 & 77.72 \\
      & mean+std   & gated  & 64.65 & 78.48 \\
      & boundary   & gated  & 64.91 & 78.30 \\
    \hline
    Wav2vec 2.0 & mean$^{\text{ext}}$ & gated & 67.00 & 77.73 \\
    \hline
  \end{tabular}
  \caption{Ablations with VSLNet on \textsc{Euronews}. The first row for each ASR model is the text-only baseline. The last line shows results on Whisper transcriptions and a dedicated audio encoder. R@1 at IoU$\geq0.5$.}
  \label{tab:ablations}
\end{table}


\begin{table*}[t]
  \centering
  \small
  \renewcommand{\arraystretch}{1.1}
  \begin{tabular}{ll|rr|rr|rr|rr}
    \hline
    \multirow{3}{*}{\textbf{Method}} &
    \multirow{3}{*}{\textbf{Text encoder}} &
    \multicolumn{4}{c|}{\textsc{Euronews}} &
    \multicolumn{4}{c}{\textsc{YTSeg}} \\
    \cline{3-10}
    & &
    \multicolumn{2}{c|}{\textbf{EM}} &
    \multicolumn{2}{c|}{\textbf{R@1}} &
    \multicolumn{2}{c|}{\textbf{EM}} &
    \multicolumn{2}{c}{\textbf{R@1}} \\
    & &
    \textbf{T} & $\boldsymbol{\Delta}_{\text{T+A}}$ &
    \textbf{T} & $\boldsymbol{\Delta}_{\text{T+A}}$ &
    \textbf{T} & $\boldsymbol{\Delta}_{\text{T+A}}$ &
    \textbf{T} & $\boldsymbol{\Delta}_{\text{T+A}}$ \\
    \hline

    \multirow{3}{*}{VSLNet}
      & MiniLM
      & 45.26 & +23.85 & 72.92 & +6.74
      & 13.88 & +6.68 & 44.69 & +3.94 \\
      & E5-large
      & 55.86 & +14.01 & 78.67 & -0.40
      & 19.07 & +3.88 & 50.42 & -2.01 \\
      & Qwen3-4B
      & 57.16 & +16.21 & 79.84 & +3.19
      & 19.68 & +5.99 & 53.25 & +2.14 \\
    \hline

    \multirow{3}{*}{QMSum-Pointer}
      & MiniLM
      & 35.34 & +26.63 & 63.18 & +11.31
      & 7.63 & +5.47 & 25.81 & +2.90 \\
      & E5-large
      & 43.11 & +20.02 & 67.40 & +8.76
      & 9.25 & +3.29 & 26.43 & +1.68 \\
      & Qwen3-4B
      & 45.49 & +18.05 & 69.24 & +5.70
      & 10.69 & +3.23 & 29.30 & +2.03 \\
    \hline
  \end{tabular}
  \caption{
    Text-encoder ablations.
    T denotes text-only performance and
    $\Delta_{\mathrm{T+A}}$ the absolute change when adding ASR-derived representations.
    R@1 is reported at IoU$\geq0.5$.
  }
  \label{tab:text-encoder-ablation}
\end{table*}

\begin{figure}[t]
  \centering
  \includegraphics[width=\columnwidth]{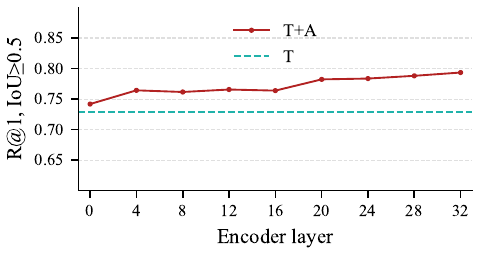}
  \caption{Performance of VSLNet, using ASR encoder states of different layers. Trained on \textsc{Euronews} with Whisper transcriptions.}
  \label{fig:layers}
\end{figure}

\begin{table*}[t]
\small
  \centering
  \renewcommand{\arraystretch}{1.1}
  \resizebox{\linewidth}{!}{
  \begin{tabular}{llll|cc|cc|cc|cc}
      \hline
      \multirow{2}{*}{\textbf{Method}} &
      \multirow{2}{*}{\textbf{Segments}} &
      \multirow{2}{*}{\textbf{Retriever}} &
      \multirow{2}{*}{\textbf{Text encoder}} &
      \multicolumn{2}{c}{\textsc{AMI}} &
      \multicolumn{2}{|c|}{\textsc{Videoaula}} &
      \multicolumn{2}{|c|}{\textsc{Euronews}} &
      \multicolumn{2}{|c}{\textsc{YTSeg}} \\
      & & & &
      \textbf{EM} & \textbf{R@1} &
      \textbf{EM} & \textbf{R@1} &
      \textbf{EM} & \textbf{R@1} &
      \textbf{EM} & \textbf{R@1} \\
    \hline

    \multirow{8}{*}{Retrieve}
       & Fixed-size & BM25  & --       & 0.00 & 6.44 & 0.13 & 9.81  & 1.62  & 40.59 & 0.28  & 19.10 \\
       & Predicted (T+A) & BM25  & --       & 0.35 & 7.50 & 8.49 & 17.64 & 72.07 & 79.16 & 19.36 & 40.36 \\
       & Fixed-size & Dense & MiniLM   & 0.05 & 6.54 & 0.27 & 10.88 & 1.62  & 38.08 & 0.27  & 13.52 \\
       & Predicted (T+A) & Dense & MiniLM   & 0.41 & 8.26 & 5.57 & 13.93 & 67.76 & 74.41 & 16.93 & 34.23 \\
       & Fixed-size & Dense & E5-large  & 0.05 & 8.06 & 0.27 & 12.47 & 1.93  & 46.48 & 0.29  & 19.20 \\
       & Predicted (T+A) & Dense & E5-large  & 0.56 & 8.72 & 6.63 & 14.72 & 73.64 & 80.96 & 17.72 & 35.92 \\
       & Fixed-size & Dense & Qwen3-4B & 0.00 & 8.01 & 0.27 & 13.26 & 1.89  & 49.03 & 0.40  & 23.60 \\
       & Predicted (T+A) & Dense & Qwen3-4B & 0.35 & 7.70 & 7.56 & 20.16 & \textbf{75.30} & 82.71 & 21.62 & 44.76 \\

    \hline
    VSLNet (T)
       & -- & -- & MiniLM
       & 0.20 & 13.18 & 4.91 & 17.37 & 42.57 & 74.00 & 13.30 & 44.74 \\
    VSLNet (T+A)
       & -- & -- & MiniLM
       & \textbf{1.04} & \textbf{15.63} & \textbf{10.48} & \textbf{21.22} & 73.82 & \textbf{83.39} & \textbf{24.94} & \textbf{49.74} \\

    \hline
    QMSum-P. (T)
       & -- & -- & MiniLM
       & 0.35 & 10.39 & 3.85 & 11.01 & 36.91 & 65.51 & 8.09 & 26.76 \\
    QMSum-P. (T+A)
       & -- & -- & MiniLM
       & 0.51 & 9.83 & 5.97 & 13.40 & 62.82 & 75.57 & 16.67 & 33.05 \\
    \hline

  \end{tabular}
  }
  \caption{
    Cross-dataset transfer results, with retrieval baselines and our topic-localization setup.
    All supervised models (segmenters and localizers) are trained on the combined train sets of \textsc{Euronews} and \textsc{YTSeg}.
    \textsc{AMI} and \textsc{Videoaula} are only used for evaluation.
    R@1 is reported at IoU$\geq0.5$.
    T=Text embeddings, A=ASR-derived embeddings, and T+A=gated fusion.
    For retrieval over predicted segments, T+A indicates that the supervised segmenter uses text and audio input; retrieval itself is text-based.
  }
  \label{tab:results-cross}
\end{table*}

\autoref{tab:ablations} reports ablations with VSLNet on \textsc{Euronews}.
We vary the ASR backbone, acoustic pooling, and fusion strategy.
The first row for each ASR model gives the text-only baseline.

\paragraph{ASR backbone.}
Text+audio improves over text-only for both ASR models, increasing EM by $23.85$ points with Whisper and $18.23$ with Canary.
This suggests that the benefit of reused ASR encoder states is not specific to one ASR model.
We also compare against a dedicated wav2vec~2.0 audio encoder~\citep{Baevski-2020}.
It performs below reused Whisper states, while requiring an additional audio encoder pass.
We therefore use Whisper for the remaining experiments.

\paragraph{Pooling and fusion.}
Among the pooling variants, Mean$^{\text{ext}}$ gives the best Whisper result and remains competitive for Canary.
For Canary, Boundary gives slightly higher EM, while Mean+Std gives slightly higher R@1.
Overall, the differences between pooling variants are small.
For fusion, gated fusion gives the best EM for both ASR models and the best or near-best R@1.
The gains over simpler fusion methods are modest, but consistent.
We therefore use Mean$^{\text{ext}}$ pooling and gated fusion in the remaining experiments.

\paragraph{Encoder layer.}
\autoref{fig:layers} shows the effect of using different Whisper encoder layers.
Performance generally improves in deeper layers, with the final layer giving the best result.
Since T+A already includes explicit text embeddings, late ASR states appear to add complementary information, potentially mixing acoustic, phonetic, lexical, and semantic signals.
This aligns with prior analyses showing that deeper speech encoder layers capture more linguistic and semantic information~\citep{Pasad-2021,Yang-Zhao-2023}.
Based on these results, we use layer $32$ for the remaining experiments.

\paragraph{Text encoder.}
To test whether the improvements depend on the relatively compact MiniLM text encoder, we additionally evaluate E5-large and Qwen3-Embedding-4B.
As shown in \autoref{tab:text-encoder-ablation}, T+A improves EM in all 12 comparisons and R@1 in 10 of 12 comparisons.
Thus, the benefits of ASR-derived representations are not limited to MiniLM, although they do not improve every metric.

\subsection{Cross-Dataset Transfer}
\label{ssec:cross-dataset-eval}

We test whether the localization models transfer beyond the datasets used for training.
For this experiment, we train VSLNet and QMSum-Pointer on the combined train sets of \textsc{Euronews} and \textsc{YTSeg}.
This combines multilingual news data with broader YouTube domains.
We evaluate on the original test sets and on two unseen datasets, \textsc{AMI} and \textsc{Videoaula}, which are not used during training.

\paragraph{Retrieval baselines.}
We compare the localization models against retrieval over either fixed-size windows or topic segments predicted by a supervised multimodal segmenter (details in \autoref{app:segment-retrieve}).
Fixed window sizes are derived from the average topic lengths in the \textsc{Euronews} and \textsc{YTSeg} dev sets and use 20\% overlap.
For retrieval, we evaluate BM25 and dense retrieval using MiniLM, E5-large, and Qwen3-Embedding-4B.

The results are shown in \autoref{tab:results-cross}.
VSLNet T+A achieves the highest R@1 on all four datasets and the highest EM on three of four.
Fixed-size windows are particularly limited under EM because their boundaries are not adaptive, but they also remain below VSLNet T+A under the more forgiving R@1 metric.
This supports query-conditioned span localization over ranking fixed windows or query-independent segments.

\begin{figure*}[t]
  \centering
  \includegraphics[width=0.8\linewidth]{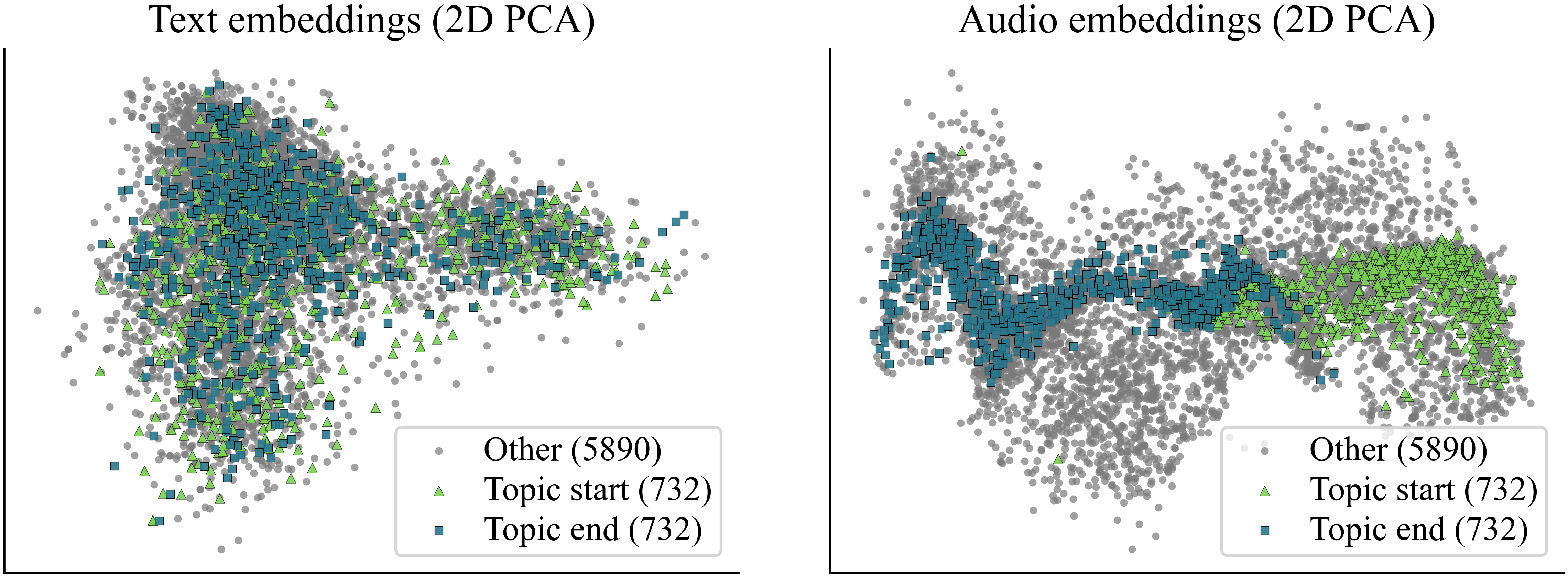}
  \caption{Visualization of sentence roles of first 100 \textsc{Euronews} dev recordings. Plots compare text embeddings and ASR-derived audio embeddings, both before projection. Dimensionality reduction using PCA lowrank.}
  \label{fig:embeddings-png}
\end{figure*}

\begin{table*}[t]
  \centering
  \small
  \renewcommand{\arraystretch}{1.05}
  \begin{tabular}{ll|cc|cc|cc|cc}
    \hline
    & & \multicolumn{2}{c|}{\textsc{Euronews}} & \multicolumn{2}{c|}{\textsc{YTSeg}} & \multicolumn{2}{c|}{\textsc{AMI}} & \multicolumn{2}{c}{\textsc{Videoaula}} \\
    \textbf{Task} & \textbf{Input} & \textbf{AUC} & \textbf{F1} & \textbf{AUC} & \textbf{F1} & \textbf{AUC} & \textbf{F1} & \textbf{AUC} & \textbf{F1} \\
    \hline
    Start    & T & 82.90 & 35.63 & 74.06 & 13.75 & 68.34 & 10.91 & 80.48 & 28.61 \\
    Start    & A & 97.76 & 70.68 & 81.42 & 18.47 & 70.15 & 11.84 & 94.42 & 48.73 \\
    \hline
    End      & T & 62.76 & 21.73 & 66.14 & 10.82 & 61.03 & 7.98 & 65.55 & 16.08 \\
    End      & A & 96.33 & 59.36 & 75.67 & 14.50 & 62.46 & 8.69 & 94.07 & 44.96 \\
    \hline
    Boundary & T & 71.69 & 42.23 & 68.63 & 20.96 & 65.76 & 16.74 & 69.18 & 33.62 \\
    Boundary & A & 95.05 & 73.59 & 78.51 & 27.92 & 68.75 & 18.02 & 90.86 & 57.14 \\
    \hline
  \end{tabular}
  \caption{Embedding probe results (in \%) for topic-boundary classification.}
  \label{tab:embedding-probe}
\end{table*}

\paragraph{Structured and semi-structured content.}
\autoref{tab:results-cross} shows that adding ASR-derived representations gives the clearest gains on the original domains and on \textsc{Videoaula}.
For VSLNet, T+A improves over text-only on all datasets.
For QMSum-Pointer, T+A improves on \textsc{Videoaula}, \textsc{Euronews}, and \textsc{YTSeg}.
This suggests that ASR states transfer beyond the training domains, especially to structured or semi-structured content.

\paragraph{Spontaneous meetings remain difficult.}
Results on \textsc{AMI} are much lower for all models.
This shows that transfer to spontaneous meetings remains challenging.
The dataset differs strongly from the training data: it contains overlapping speech, multiple speakers, and less structured topic transitions.
Audio still improves VSLNet on \textsc{AMI}, but the gains are weak; for QMSum-Pointer, T+A slightly improves EM but reduces R@1.
Overall, cross-dataset transfer depends strongly on the target domain.
We qualitatively inspect acoustic cues that may help explain this pattern in \autoref{sec:analysis}.


\section{Analysis}
\label{sec:analysis}
We provide analyses to help explain the observed localization gains.
First, we test whether text and ASR-derived embeddings encode boundary roles.
Second, we qualitatively inspect acoustic cues at topic boundaries.

\subsection{Embedding Analysis}
\label{ssec:embedding-analysis}
The main results suggest that ASR-derived embeddings are especially useful for boundary precision.
We therefore test whether sentence embeddings encode topic-boundary roles.
For each sentence, we define three probe tasks: \emph{start} (topic start vs.\ other), \emph{end} (topic end vs.\ other), and \emph{boundary} (start-or-end vs.\ other).
We train linear probes on text embeddings (T) and ASR-derived audio embeddings (A), and report AUC-ROC and F1 in \autoref{tab:embedding-probe}.

Across all datasets and probe tasks, ASR-derived audio embeddings are more predictive of boundary roles than text embeddings.
The largest differences appear for \textsc{Euronews} and \textsc{Videoaula}, while gains are smaller for \textsc{AMI}.
This is consistent with the localization results, where adding ASR-derived representations yields the largest gains on structured or semi-structured content.
The PCA visualization in \autoref{fig:embeddings-png} shows the same trend for \textsc{Euronews}: text embeddings overlap strongly across sentence roles, while ASR-derived embeddings form clearer boundary regions.
Overall, these results show that ASR-derived embeddings encode boundary-sensitive information, consistent with the larger gains under stricter boundary-matching criteria.
However, the probes do not determine which acoustic or linguistic information in the ASR states drives the gains.

\newcommand{\cuedot}{\(\bullet\)}
\newcommand{\cuescore}[1]{%
  \ifcase#1
    --%
  \or
    \cuedot%
  \or
    \cuedot\hspace{0.1em}\cuedot%
  \or
    \cuedot\hspace{0.1em}\cuedot\hspace{0.1em}\cuedot%
  \or
    \cuedot\hspace{0.1em}\cuedot\hspace{0.1em}\cuedot\hspace{0.1em}\cuedot%
  \or
    \cuedot\hspace{0.1em}\cuedot\hspace{0.1em}\cuedot\hspace{0.1em}\cuedot\hspace{0.1em}\cuedot%
  \fi
}


\begin{table}[t]
\centering
\small
\renewcommand{\arraystretch}{1.1}
\resizebox{\linewidth}{!}{
\begin{tabular}{llcccc}
\hline
\textbf{Cue type} & \textbf{Aspect}
& \textsc{Eur} & \textsc{YTS} & \textsc{Vid} & \textsc{AMI} \\
\hline

\multirow{2}{*}{End prosody}
 & Pres.
 & \cuescore{3} & \cuescore{2} & \cuescore{4} & \cuescore{1} \\
 & Dist.
 & \cuescore{3} & \cuescore{2} & \cuescore{2} & \cuescore{1} \\
\hline

\multirow{2}{*}{Start prosody}
 & Pres.
 & \cuescore{2} & \cuescore{2} & \cuescore{4} & \cuescore{1} \\
 & Dist.
 & \cuescore{2} & \cuescore{2} & \cuescore{2} & \cuescore{1} \\
\hline

\multirow{2}{*}{Discourse markers}
 & Pres.
 & \cuescore{1} & \cuescore{2} & \cuescore{3} & \cuescore{1} \\
 & Dist.
 & \cuescore{1} & \cuescore{3} & \cuescore{4} & \cuescore{1} \\
\hline

\multirow{2}{*}{Pauses}
 & Pres.
 & \cuescore{4} & \cuescore{3} & \cuescore{4} & \cuescore{2} \\
 & Dist.
 & \cuescore{3} & \cuescore{2} & \cuescore{4} & \cuescore{1} \\
\hline

\multirow{2}{*}{Speaker/language}
 & Pres.
 & \cuescore{4} & \cuescore{1} & \cuescore{0} & \cuescore{2} \\
 & Dist.
 & \cuescore{3} & \cuescore{1} & \cuescore{0} & \cuescore{1} \\
\hline

\multirow{2}{*}{Production cues}
 & Pres.
 & \cuescore{5} & \cuescore{2} & \cuescore{0} & \cuescore{0} \\
 & Dist.
 & \cuescore{5} & \cuescore{3} & \cuescore{0} & \cuescore{0} \\

\hline
\end{tabular}
}
\caption{
Qualitative single-author assessment of cues observed around topic boundaries.
Presence (Pres.) indicates how frequently a cue was observed, while distinctiveness (Dist.) indicates how clearly it differentiated topic boundaries from regular sentence transitions.
Scores range from 0--5, shown as -- to \cuescore{5}.
}
\label{tab:cue-analysis}
\end{table}

\subsection{Acoustic Cue Analysis}
\label{ssec:cue-analysis}

To better understand which acoustic signals may support localization, we manually inspected topic-boundary contexts.
For each dataset, one author randomly sampled 20 recordings and listened to the topic-shift contexts within these recordings (more details in \autoref{app:cue-analysis}).
We grouped the observed cues into six categories and separately assessed their presence and boundary distinctiveness.
The latter is a subjective qualitative rating indicating how clearly a cue distinguishes topic boundaries from regular sentence boundaries.

\autoref{tab:cue-analysis} shows clear domain differences.
\textsc{Euronews} has the most distinct boundary cues, especially production cues such as swoosh sounds and background changes, as well as pauses and speaker changes.
This is consistent with the strong audio-only and text+audio results on this dataset.
\textsc{YTSeg} is more heterogeneous: tutorials and reviews often contain pauses, discourse markers, or sound effects, while lectures tend to have weaker cues.
\textsc{Videoaula} contains no production cues, but topic boundaries are often accompanied by pauses and vocal patterns, such as falling intonation at topic ends and higher pitch or emphasis at topic starts.
In contrast, \textsc{AMI} contains few boundary-specific cues.
Pauses and speaker changes occur around some topic boundaries, but are also common elsewhere in the recordings, while topic shifts are often gradual and interaction-driven.

Overall, the observed cue patterns provide a plausible explanation for the domain differences in localization performance.
Recurring acoustic and discourse-structural boundary cues are clearest in structured news content, still present in prepared lectures and tutorials, and weakest in spontaneous meetings.
This is consistent with the cross-dataset results, where adding ASR-derived representations yields larger gains on \textsc{Videoaula} than on \textsc{AMI}.
Addressing this gap is a useful direction for future work, as both text-only and audio-augmented localizers struggle in this domain.

\section{Conclusion}
We studied query-conditioned topic localization in speech transcripts, where the goal is to find the sentence span that best addresses a textual query.
To add speech information to this task, we reused ASR encoder states as sentence-level acoustic representations and fused them with textual embeddings.
This allows span locators to use ASR-derived information without running a separate audio encoder.

Our experiments show that text+audio models improve over text-only baselines across the main in-domain settings and across two localization architectures.
The gains are strongest under stricter boundary-matching criteria, indicating that reused ASR representations provide complementary information for refining start and end positions.  
Cross-dataset results further indicate that the benefits are strongest for structured or semi-structured spoken content, while spontaneous meetings remain challenging.

\section*{Limitations}
Our work focuses on intra-document topic localization: the relevant spoken document is assumed to be known, and the model predicts the query-relevant span within that document.
This reflects a common second-stage retrieval setting, where document retrieval is performed first and within-document localization is applied only to candidate recordings.
Combining both stages in an end-to-end retrieval pipeline is left to future work.

We use topic titles as queries.
These titles can be viewed as concise, normalized descriptions of the target information need, similar to rewritten queries in practical retrieval systems.
However, real user queries may be noisier, more ambiguous, or less specific.
To assess sensitivity to this setup, we include a controlled query-perturbation experiment in \autoref{app:query-perturbation}, where titles are replaced by natural-language templates or single-typo variants.
All perturbations reduce performance, with the largest degradation for typo queries, indicating that robustness to more realistic user queries remains an open challenge.

Our models predict a single contiguous span, matching the annotation structure of the datasets and the goal of retrieving one focused context window.
Some information needs may require multiple spans or evidence distributed across a recording.
Extending the method to multi-span localization is an important direction for future work.

The cross-dataset evaluation is constrained by the availability of annotated spoken-topic datasets with audio, boundaries, and titles.
While \textsc{AMI} and \textsc{Videoaula} allow us to test transfer to meetings and lectures, they are much smaller than the main training datasets.
Moreover, results on \textsc{AMI} show that generalization to spontaneous multi-speaker meetings remains limited. 
Larger and more diverse benchmarks would support a more complete assessment of cross-domain generalization.

Our study focuses on reusing ASR representations to augment text-based localization and does not incorporate visual information.
Visual streams are also not consistently available across the datasets considered.
Extending the approach with video representations is an interesting direction for future work.

\section*{Ethical Considerations}
Our work processes speech recordings and derives sentence-level representations from ASR encoder states.
Such representations may contain information beyond the spoken words, including speaker identity, accent, speaking style, emotion, or recording conditions~\citep{Ruggiero-2025}.
This is relevant from a privacy perspective, as prior work has shown that self-supervised speech models can leak membership information at both the utterance and speaker level~\citep{Tseng-2022}.
In our proposed pipeline, however, the stored representations are projected and fused with text embeddings for topic localization.
This may reduce speaker-specific information compared to storing raw ASR encoder states, but does not eliminate privacy risks.
They should therefore be treated as potentially sensitive data.
In practical deployments, ASR-derived embeddings should be stored securely, access should be restricted, and embeddings should not be made publicly available when the underlying recordings contain personal or private speech.


\bibliography{custom}

\appendix

\section{Dataset Statistics}
\label{app:dataset-statistics}
\autoref{tab:datastats} shows an overview of the datasets used in the experiments (\autoref{sec:experiments}).
\autoref{tab:flops} shows an overview of the FLOPs needed for an average forward pass on the samples of the \textsc{Euronews} test set. 
In addition to the numbers shown in \autoref{tab:main-results}, this table breaks down the numbers into the different modules of the offline and online stage.
Further, it comprises the FLOPs needed by the text embedding model, which is excluded from the localization-stage FLOPs reported in \autoref{tab:main-results}.
ASR transcription itself is not included.

\begin{table*}[h]
\small
  \centering
  \renewcommand{\arraystretch}{1.2}
  \begin{tabular}{llll|r|cc|c}
    \hline
      \multirow{2}{*}{\textbf{Dataset}}  &  \multirow{2}{*}{\textbf{Lang}} &  \multirow{2}{*}{\textbf{Domain}}  & \multirow{2}{*}{\textbf{Partition}} &  \multirow{2}{*}{\textbf{\# Rec}} & \multicolumn{2}{c|}{\textbf{Duration mm:ss}} & \multirow{2}{*}{\textbf{Topics/Rec}} \\
     &  &  &  &  & Recording & Topic &  \\
    \hline
    \multirow{3}{*}{\textsc{Euronews}}  & \multirow{3}{*}{multi} & \multirow{3}{*}{news}  & train & 4964 & 10:56 $\pm$ 5:07 & 1:29 $\pm$ 1:06 & 7.17 $\pm$ 2.46 \\
       &  &   & dev &   276 & 10:46 $\pm$ 5:01 & 1:22 $\pm$ 0:45 & 7.81 $\pm$ 3.25 \\
       &  &   & test & 277 & 11:34 $\pm$ 4:23 & 1:25 $\pm$ 0:49 & 10.65 $\pm$ 6.65 \\
    \hline
    \multirow{3}{*}{\textsc{YTSeg}}  & \multirow{3}{*}{en}    & \multirow{3}{*}{mixed} & train & 16403 & 20:16 $\pm$ 25:15 & 2:17 $\pm$ 3:20 & 8.75 $\pm$ 5.36 \\
         &    &  & dev  & 1443 & 20:43 $\pm$ 25:09 & 2:18 $\pm$ 3:20 & 8.88 $\pm$ 5.67 \\
         &    &  & test & 1448 & 20:40 $\pm$ 26:24 & 2:18 $\pm$ 3:24 & 8.81 $\pm$ 5.74 \\
    \hline
    \textsc{AMI}        & en    & meetings & test & 138 & 31:37 $\pm$ 9:59 & 3:05 $\pm$ 3:17 & 14.30 $\pm$ 7.61  \\
    \hline
    \textsc{Videoaula}  & pt    & lectures & test  & 34 & 40:52 $\pm$ 20:00 & 3:38 $\pm$ 6:22 & 22.18 $\pm$ 8.93 \\
    \hline
  \end{tabular}
  \caption{Dataset statistics; reported $mean \pm stddev$; multi=\{en, de, es, fr, pt, it\}.}
  \label{tab:datastats} 
\end{table*}

\begin{table*}[t]
  \centering
  \small
  \renewcommand{\arraystretch}{1.2}
  \begin{tabular}{ll|rrr} 
    \hline
    \multirow{2}{*}{\textbf{Stage}} & \multirow{2}{*}{\textbf{Module}} & \multicolumn{3}{c}{\textbf{Input}} \\
          &        & T & A & T+A \\
    \hline
    \multirow{5}{*}{Offline} & ASR pooling  & - & 57,086,045 & 57,086,045 \\
                             & Text emb     & 274,575,236,920 & 274,575,236,920 & 274,575,236,920 \\
                             & Index        & 8,502,409 & 28,341,363 & 45,346,180 \\
                             & Context enc  & 117,749,276 & 117,749,276 & 117,749,276  \\
                             \cline{2-5}
                             & TOTAL        & 274,701,488,605 & 274,778,413,604 & 274,795,418,421 \\
    \hline
    \multirow{5}{*}{Online}  & Text emb  & 450,572,607 & 450,572,607 & 450,572,607 \\
                             & VSLNet    &  37,234,682 & 37,234,682 & 37,234,682 \\
                             & QMSum     & 21,202,532 & 21,202,532 & 21,202,532 \\
                             \cline{2-5}
                             & TOTAL VSLNet   & 487,807,289 & 487,807,289 & 487,807,289  \\
                             & TOTAL QMSum    & 471,775,139 & 471,775,139 & 471,775,139 \\
    \hline
  \end{tabular}
  \caption{FLOPs by stage, module, localizer and input modalities. Values are averaged over \textsc{Euronews} test samples.}
  \label{tab:flops}
\end{table*}

\section{Retrieval Baselines}
\label{app:segment-retrieve}


The retrieval baselines rank candidate transcript segments according to their relevance to the query.
We consider two strategies for segmenting the transcript: (i) using fixed-size windows and (ii) predicting topic segment boundaries.

For predicting segments, we use the supervised multimodal topic segmenter from \citet{Freisinger-2026}.
It consists of a pretrained text encoder, a pretrained wav2vec~2.0 audio encoder, and a transformer-based tagger that predicts binary topic-change labels at sentence level.
The segmenter is trained on the combined \textsc{Euronews} and \textsc{YTSeg} training sets.

As a simpler alternative, we split transcripts into fixed-size windows with 20\% overlap.
The window sizes are derived from the average topic-segment lengths in the \textsc{Euronews} and \textsc{YTSeg} development sets to avoid tuning on the target test datasets.

For retrieval, we evaluate BM25 and dense retrieval.
For dense retrieval, candidate spans and queries are embedded using MiniLM, E5-large, or Qwen3-Embedding-4B, and ranked by cosine similarity.
The highest-ranked candidate span is used as the final prediction.

\section{Additional Analysis Details}
\label{app:analysis-details}

\subsection{Embedding Probe Details}
\label{app:embedding-analysis}
For the embedding probes, we use frozen sentence-level embeddings as input and train one linear classifier per task.
The three binary tasks are \emph{start} (topic start vs.\ other), \emph{end} (topic end vs.\ other), and \emph{boundary} (start-or-end vs.\ other).
We train probes separately for text embeddings (T) and ASR-derived audio embeddings (A).
For \textsc{Euronews} and \textsc{YTSeg}, probes are trained on the train split and evaluated on the dev split.
For \textsc{AMI} and \textsc{Videoaula}, we use an 80/20 train/eval split.
All probes are trained with weighted binary cross-entropy to account for class imbalance.
We use AdamW optimizer, a learning rate of 1e-2, a batch size of 4096 and train for 200 epochs.
We report AUC-ROC and F1 on the evaluation split.
The PCA visualization in \autoref{fig:embeddings-png} is used only for qualitative inspection and is not used for probe training or evaluation.

\section{Acoustic Cue Analysis Details}
\label{app:cue-analysis}
We provide additional details for the manual cue analysis in \autoref{ssec:cue-analysis}.
For each dataset, one author randomly sampled 20 recordings and inspected topic-boundary contexts and regular sentence boundary contexts using a $\pm 5$s window around boundaries.
For \textsc{Euronews} and \textsc{YTSeg}, recordings are sampled from the dev sets.

The analysis was used to identify recurring cue types and to qualitatively characterize their presence and boundary distinctiveness.
Cue presence describes whether and to what extent a cue was observed in the inspected boundary contexts, whereas boundary distinctiveness describes how clearly the cue distinguished topic boundaries from regular sentence transitions.
Both dimensions are rated on a qualitative scale from 0--5.
For presence, higher scores indicate that a cue was observed more frequently around topic boundaries.
For distinctiveness, higher scores indicate that a cue more clearly differentiates topic boundaries from regular sentence transitions.
The assessment is subjective and was performed by a single author.
These observations are intended to support interpretation of the localization results and should not be interpreted as evidence that the localization models encode or use the identified cues.
\autoref{tab:cue-definitions} defines the cue categories, and \autoref{tab:cue-dataset-observations} summarizes the main dataset-level observations.

\begin{table*}[t]
\centering
\small
\setlength{\tabcolsep}{5pt}
\renewcommand{\arraystretch}{1.2}
\resizebox{\linewidth}{!}{
\begin{tabular}{p{0.14\linewidth}p{0.25\linewidth}p{0.30\linewidth}p{0.23\linewidth}}
\hline
\textbf{Dataset} & \textbf{Speech/content style} & \textbf{Typical observed cues} & \textbf{Interpretation} \\
\hline
\textsc{Euronews} &
Structured broadcast news. &
Frequent transition sounds, background-noise changes, speaker changes, pauses, occasional language changes or foreign-language excerpts, falling intonation at topic ends. &
Cues are highly distinct and often production-driven. This is consistent with the strong audio-only and text+audio results. \\
\hline
\textsc{YTSeg} &
Heterogeneous YouTube content, including tutorials, lectures, reviews, interviews, news, and informative videos. &
Pauses, pitch resets, emphatic starts, discourse markers such as "So", "Now", "Let's", occasional swoosh effects, music changes, background changes, and explicit numbering. &
Cues are present but vary strongly by format. Tutorials and reviews are often clearer than lectures or freer speech. \\
\hline
\textsc{Videoaula} &
Portuguese computer-science lectures with guided, semi-spontaneous speech. &
Pauses, falling intonation at topic ends, rising pitch or increased volume at topic starts, and markers such as "Bem" or "Bom". No production cues or speaker changes. &
Boundaries are mainly vocal and discourse-structural. Yet, the cues are less explicit than in edited news. \\
\hline
\textsc{AMI} &
Spontaneous multi-speaker meetings. &
Frequent speaker changes, pauses, hesitations, incomplete speech, laughter, and markers such as "Okay", "Right", "Well", or "So". &
Many cues occur throughout the meeting and are not boundary-specific. Topic changes are often gradual and tied to interaction shifts. \\
\hline
\end{tabular}
}
\caption{Dataset-level observations from the manual acoustic cue analysis.}
\label{tab:cue-dataset-observations}
\end{table*}

\begin{table}[t]
\centering
\small
\setlength{\tabcolsep}{6pt}
\renewcommand{\arraystretch}{1.2}
\resizebox{\columnwidth}{!}{
\begin{tabular}{p{0.25\linewidth}p{0.68\linewidth}}
\hline
\textbf{Cue type} & \textbf{Examples} \\
\hline
End prosody &
Falling intonation, reduced energy, lower pitch, slower speech rate, lowered voice at topic end. \\
Start prosody &
Higher pitch, emphatic or energetic start, increased volume, reset of voice, faster speech rate. \\
Discourse markers &
Initial words such as "So", "Now", "Okay", "Right", "Well", Portuguese "Bem"/"Bom", or explicit numbering such as "Number 3". \\
Pauses &
Short or medium silence between topic end and next topic start. \\
Speaker/language &
Speaker change/alternation, language change, foreign-language excerpt. \\
Production cues &
Swoosh or transition sound, jingle, background music, music swell, change in background noise, inserted recording excerpt, scene or sound-setting change. \\
\hline
\end{tabular}
}
\caption{Cue categories used in the manual acoustic cue analysis.}
\label{tab:cue-definitions}
\end{table}


\begin{table}[t]
    \centering
    \small
    \renewcommand{\arraystretch}{1.2}
    \resizebox{\linewidth}{!}{
    \begin{tabular}{ll|cc|cc}
    \hline
    \textbf{Query} & \textbf{Input} & \multicolumn{2}{c|}{\textsc{Euronews}} & \multicolumn{2}{c}{\textsc{YTSeg}} \\
     & & ($\Delta$)EM & ($\Delta$)R@1 & ($\Delta$)EM & ($\Delta$)R@1 \\
    \hline
    \multirow{2}{*}{Title} 
        & T   & 45.26 & 72.92 & 13.88 & 44.69 \\
        & T+A & 69.11 & 79.66 & 20.56 & 48.63 \\
    \hline
    \multirow{2}{*}{Short} 
        & T   & -4.58 & -5.83 & -1.77 & -5.75 \\
        & T+A & -8.94 & -9.66 & -2.30 & -5.08 \\
    \hline
    \multirow{2}{*}{Long} 
        & T   & -5.25 & -7.63 & -2.49 & -8.19 \\
        & T+A & -8.18 & -9.48 & -3.45 & -8.29 \\
    \hline
    \multirow{2}{*}{Typo} 
        & T   & -7.36 & -12.03 & -3.68 & -10.55 \\
        & T+A & -11.23 & -13.65 & -5.41 & -11.41 \\
    \hline
    \end{tabular}
    }
    \caption{
        Query perturbation results for VSLNet.
        Title rows show absolute EM and R@1 scores at IoU$\geq0.5$.
        All other rows show changes relative to the Title setting.
        Negative values indicate performance drops.    
    }
    \label{tab:perturb}
\end{table}

\section{Query Perturbation Analysis}
\label{app:query-perturbation}
Our main experiments use topic titles as queries.
This is a controlled setup, but real user queries may be longer, less direct, or contain spelling errors.
We therefore evaluate how sensitive the VSLNet locator is to simple query perturbations.
We keep the trained models from \autoref{ssec:expmain} fixed and modify only the test queries.

We compare the original title query with three perturbations.
\textit{Short} wraps the title in a short natural language query, such as "Find the part about \{title\}." 
\textit{Long} uses a longer template, such as "I am looking for information about \{title\}." 
\textit{Typo} inserts one spelling error into the title.
For \textsc{Euronews}, templates are translated into the language of the corresponding recording.
We evaluate both text-only and text+audio VSLNet models on \textsc{Euronews} and \textsc{YTSeg}.

\autoref{tab:perturb} reports the change in EM and R@1 relative to the original title query.
All perturbations reduce performance, with the strongest drops for typo queries.
This indicates that title-based queries are easier than more natural or noisy queries.
However, text+audio remains sensitive to the same perturbations as text-only input, suggesting that audio does not remove the need for robust query handling.
Future work should therefore consider query rewriting, training with natural user queries, or query augmentation.

\end{document}